\documentclass[letterpaper, 10 pt, journal, twoside]{IEEEtran}

\usepackage{graphicx} 
\usepackage{amsmath} 
\usepackage{amssymb}  
\usepackage{soul}
\usepackage{algorithm}
\usepackage[noend]{algpseudocode}
\usepackage{wrapfig, blindtext}
\usepackage{multirow}

\usepackage[font=small,skip=2pt]{caption}

\newcommand{\truestatespace}{\mathcal{X}}
\newcommand{\controlspace}{\mathcal{U}}
\newcommand{\observationspace}{\mathcal{O}}
\newcommand{\latentstatespace}{\mathcal{Z}}
\newcommand{\cost}{c}
\newcommand{\truestate}{x}
\newcommand{\control}{u}
\newcommand{\observation}{o}

\newcommand{\truedynamics}{f}
\newcommand{\observationfn}{g}

\newcommand{\latentobservationfn}{\hat{g}}

\newcommand{\policy}{\pi}
\newcommand{\latentstate}{z}
\newcommand{\latentobservation}{y}
\newcommand{\encoder}{\phi}
\newcommand{\parameters}{\rho}
\newcommand{\methodname}{LVSPC}
\renewcommand{\algorithmicindent}{1.2em}

\newcommand{\latentdynamics}{\hat{f}_{\parameters}}

\newcommand\Algphase[1]{%
\vspace*{-.7\baselineskip}\Statex\hspace*{\dimexpr-\algorithmicindent-2pt\relax}\rule{0.47\textwidth}{0.4pt}%
\Statex\hspace*{-\algorithmicindent}\textbf{#1}%
\vspace*{-.7\baselineskip}\Statex\hspace*{\dimexpr-\algorithmicindent-2pt\relax}\rule{0.47\textwidth}{0.4pt}%
}
\begin{document}

\title{Keep it Simple: Data-efficient Learning for Controlling Complex Systems with Simple Models}

\author{Thomas Power$^1$ and Dmitry Berenson$^1$
\thanks{Manuscript Received: October 16, 2020; Revised: December 21, 2020; Accepted: January 15, 2021.}%
\thanks{This paper was recommended for publication by Editor Dana Kulic upon evaluation of the Associate Editor and Reviewers’ comments. This work was supported in part by NSF Grant IIS-1750489 and ONR grant N00014-21-1-2118.}%
\thanks{$^1$Authors are with the University of Michigan, Ann Arbor, MI, USA. {\tt\small \{tpower, dmitryb\}@umich.edu}}%
\thanks{
Digital Object
Identifier (DOI): see top of this page
}%
}
\markboth{IEEE Robotics and Automation Letters. Preprint Version. Accepted January, 2021}
{Power and Berenson: Data-efficient Learning for Controlling Complex Systems with Simple Models} 

\maketitle

\begin{abstract}

When manipulating a novel object with complex dynamics, a state representation is not always available, for example for deformable objects. Learning both a representation and dynamics from observations requires large amounts of data. We propose Learned Visual Similarity Predictive Control (\methodname{}), a novel method for data-efficient learning to control systems with complex dynamics and high-dimensional state spaces from images. \methodname{}  leverages a given simple model approximation from which image observations can be generated. We use these images to train a perception model that estimates the simple model state from observations of the complex system online. We then use data from the complex system to fit the parameters of the simple model and learn where this model is inaccurate, also online. Finally, we use Model Predictive Control and bias the controller away from regions where the simple model is inaccurate and thus where the controller is less reliable. We evaluate \methodname{}  on two tasks; manipulating a tethered mass and a rope.  We find that our method performs comparably to state-of-the-art reinforcement learning methods with an \textit{order of magnitude less data}. \methodname{} also completes the rope manipulation task on a real robot with 80\% success rate after only 10 trials, despite using a perception system trained only on images from simulation.

\end{abstract}
\section{Introduction}
\begin{figure*}[h]
    \centering
    \includegraphics[width=0.99\textwidth]{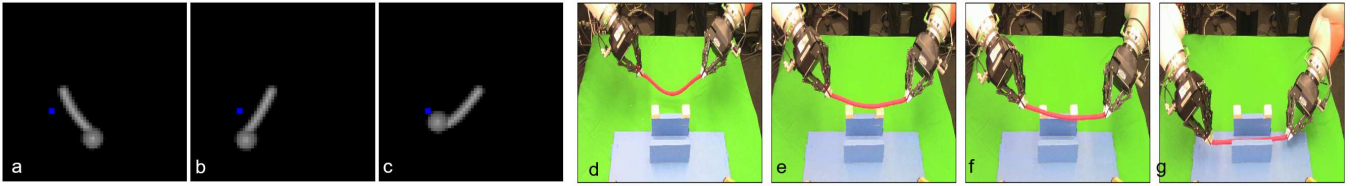}
    \caption{\textit{(a-c)}: \methodname{} controlling a tethered mass to a desired position (blue) from images by treating it as a cart-pole; \textit{(d-g)}: \methodname{} brings a rope to a target location in a narrow passage between two obstacles while avoiding protrusions by treating the rope as a rigid object. The robot starts with the rope slack but pulls it taut to keep the approximation more accurate, allowing it to complete the task.}
    \label{fig:introfig}
    \vspace{-0.5cm}
\end{figure*}
\IEEEPARstart{W}{hile} recent machine learning methods have been effective for many manipulation tasks, they rely on access to large datasets of the system being manipulated \cite{FinnRSSVF, VisualPlanningActingWang, PokebyPoking}. Yet in many scenarios we do not have time to gather extensive training data with an object before performing a task. Sim-to-real transfer has been used to fine-tune parameters on limited real-world data when the real object is similar to those used in simulation \cite{James2017Sim2Real, Chebotar2019Sim2Real}, but these methods struggle if the objects are significantly different. We would like to use prior knowledge about the object to reduce the data required for learning, but the question of \textit{how} to effectively use prior knowledge when encountering a \textit{novel} object remains open.

This paper addresses how to leverage dynamics models of simple systems when learning to control much more complex, but related, systems online. While it is possible to learn dynamics using only online data (e.g. \cite{PlaNet}), we wish to use our knowledge of a simple model to make the learning much more data-efficient, and thus practical for real-world application. For example, consider a tethered mass being swung by a gripper (Figure \ref{fig:introfig}). The dynamics of the system are complex and require a great deal of data to learn. However, if we treat the system as a cart with a rigid pendulum, we can predict the dynamics fairly accurately \textit{for some subset of the state-action space}. We can exploit this subset to perform tasks such as bringing the mass to a target, even without a globally-accurate dynamics model. Simple models are often used in this way, for example in deformable object manipulation \cite{MillerGFolding2012,WhenToTrustModelMcconachie} and control for humanoids \cite{SimpleModelsLocomotion}. 

To use knowledge of the dynamics of the simple model to control the more complex true system, we must know which states of the complex system correspond to which states of the simple system. What makes this problem especially difficult is that, while we can design a useful state representation for the simple system offline, we do not know what state representation to use for the complex system, so we cannot explicitly define a correspondence between states.

Our key insight for overcoming this problem is that the simple system (and its state representation) is a good approximation of the complex system when it gives rise to similar image observations to the complex system. By using a metric for observation similarity that reasons about uncertainty we can build a controller for the complex system and also learn where our approximation is inaccurate (to avoid visiting those parts of the state space). By utilizing domain randomization during training, we enable a single simple system state to elicit a wide variety of image observations; i.e. shapes, colors, and obstacles can vary while still producing an image we consider to be \textit{visually-similar}. We use online system identification to estimate the parameters of the simple model, however, deciding which class of simple model to use for a given task is not within the scope of this paper. Here we made this decision manually but seek to automate selecting the class of simple model in future work.

This paper makes the following contributions: 1) Learned Visual Similarity Predictive Control (\methodname{} \hspace{-3pt}), a novel framework for learning how to perform manipulation tasks with a complex system given only a simple model and images from a small number of trials online; 2) Evaluation of \methodname{} on manipulating a tethered mass (using a cart-pole as a simple model) and a rope (using a rigid body as a simple model) (See Fig. \ref{fig:introfig}) in simulation, showing large improvements in data-efficiency over baselines (PlaNet \cite{PlaNet} and CURL \cite{CURL}).  \methodname{} also completes the rope manipulation task \textit{on a real robot} with 80\% success rate after only 10 trials.

\methodname{} consists of two phases: 1) Offline, we train an ensemble Convolutional Neural Network (CNN) perception system on image observations of the simple system, outputting an estimate of the simple system's state. 2) Online, given image observations of the complex system, we do system identification to estimate parameters of the simple system dynamics and learn a Gaussian Process (GP) that predicts where the simple model is accurate. We use the simple model and the GP to track the object via a Gaussian Process Unscented Kalman Filter (GPUKF) \cite{GPUKF} and perform control via Model Predictive Path Integral Control (MPPI) \cite{MPPI}, biasing the system away from inaccurate transitions. 


\section{Related Work}
\textit{Dynamics from Images}: Learning-based approaches using dynamics models for control with images observations have included learning dynamics models directly in image space \cite{FinnRSSVF, PokebyPoking, DeepVF}. Dynamics in image space are highly complex, and these methods require large amounts of data. Other methods learn dynamics in a lower-dimensional latent space \cite{RCE, VisualPlanningActingWang, PlaNet, Dreamer}. None of these methods incorporate prior knowledge. SE3-PoseNets \cite{SE3PoseNetsByravan} learn dynamics in pose-space from point cloud data. \cite{StateEstimationDeformableObjectManipulation} use the positions of a set of ordered points as the representation of a rope and pre-trains a state estimator on ground truth in a simulator. 
Unlike \methodname{}, neither of these methods use a given model approximation nor do they reason about model uncertainty.

\textit{Using simplified models}: Simplified models have been widely explored in the legged robotics literature, in particular using spring-mass damper models \cite{SimpleModelsAtlas, SimpleModelsLocomotion}. Simplified models have been used to generate trajectories for a lower-level controller to track with guarantees \cite{RTD}. However, these guarantees require access to a high-fidelity model. Other work \cite{BanditsMcconachie} has used a set of simple models and a selection mechanism to choose between them. \cite{WhenToTrustModelMcconachie} use a given simplified dynamics model and learns a classifier on whether a given transition is reliable. We use GP uncertainty to model transition reliability rather than a classifier. We also use image observations and perform tracking concurrently.

\textit{Incorporating model uncertainty}: Previous work has shown that reasoning about model uncertainty can improve data efficiency \cite{DeisenrothPILCO, PETSChua}. PILCO \cite{DeisenrothPILCO} uses a Gaussian Process dynamics model for model uncertainty and achieves high data efficiency on learning control policies. Gaussian Processes dynamics have also been used for the purpose of both avoiding uncertainty \cite{RiskSensitiveiLQR}, or explicitly seeking it \cite{CuriousiLQR}. PETS \cite{PETSChua} uses a probabilistic ensemble of neural networks to model uncertainty and is able to outperform PILCO on control tasks with high state dimension. These methods have only been demonstrated on tasks for which state is available, and not on image domains where parameterizing uncertainty can be difficult. \methodname{} aims to combine modeling of uncertainty in the dynamics with strong priors to maintain high data efficiency when learning from images.
\section{Problem Statement}
We consider a nonlinear discrete-time system with state $\truestate \in \truestatespace$ and controls $\control \in \controlspace$. The system has unknown true dynamics given by $\truestate_{t+1} = \truedynamics(\truestate_t, \control_t)$.
We assume $\truestatespace$ may be arbitrarily high-dimensional and unobserved. Instead we may only have access to observations $\observation \in \observationspace$ via an observation function at the current state $\observation_t = \observationfn(\truestate_t)$.


We define a trial as a time-limited attempt to find a sequence of controls $\{\control_1, ..., \control_T\}$ such that the final state $x_T \in \truestatespace_{goal}$ where $\truestatespace_{goal}$ is the trial's goal region. We assume that we can fully observe when the system has reached the goal i.e. $\observation \in \observationspace_{goal} \iff \truestate \in \truestatespace_{goal}$. The goal in observation space is defined as $\observationspace_{goal} = \{\observationfn(\truestate) : \truestate \in \truestatespace_{goal}\}$. We assume that data collection on the true system is expensive. 
The unknown dynamics and high-dimensional state make this problem intractable to solve with a small dataset. Instead we seek to model the system in a latent state of lower dimensionality $\latentstate \in \latentstatespace$ with \textit{simple} dynamics $\latentdynamics$ parameterized by $\parameters$ with input-dependent noise. The transition distribution, which we will denote as $p_z$ for shorthand is given by
\begin{equation}
    p(\latentstate_{t+1} | \latentstate_t, \control_t) = \mathcal{N}(\latentdynamics(\latentstate_t, \control_t), Q(\latentstate_t, \control_t))
    \label{eq:dynamics}
\end{equation}
We assume that $\latentdynamics$ is given and is differentiable with respect to $(\latentstate, \control, \parameters)$. $Q$ is an input-dependent uncertainty term. We also assume that the simple dynamics are Markovian. The simple system has the same observation space $\observationspace$ and has a given observation function $\observation_t = \latentobservationfn(\latentstate_t)$.
We assume that we can \textit{a priori} specify some subset of the goal region in $\latentstatespace$ as $\latentstatespace_{goal}$, i.e that $\{\latentobservationfn(\latentstate) : \latentstate \in \latentstatespace_{goal}\} \subset \observationspace_{goal}$. This could also be done by specifying $\observationspace_{goal}$ directly (as is common in learning to control from images, e.g. \cite{visualforesight}) and using this to infer $z_{goal}$. 
We then seek to design a feedback policy $\control_t = \policy(\latentstate_t)$ such that $\latentstate_T \in \latentstatespace_{goal}$ for some time $T$.  Our goal is to design $\policy$ using $\latentdynamics$ so that it achieves high success rate after a small number of trials.
\section{Methods}

Our approach to this problem requires input in the form of a simple model approximation that is believed to accurately represent the dynamics over some subset of the complex system $(\truestatespace, \controlspace$). By using this simple model in simulation we can generate large amounts of data. The key to our approach is to leverage this data and our knowledge of the simple system. We then reduce the problem of unsupervised representation and dynamics learning to that of supervised learning of a perception system for the simple model representation (offline), and then learning when this representation and the dynamics are accurate (online).

\begin{figure*}[h]
     \centering
     \includegraphics[width=.99\textwidth]{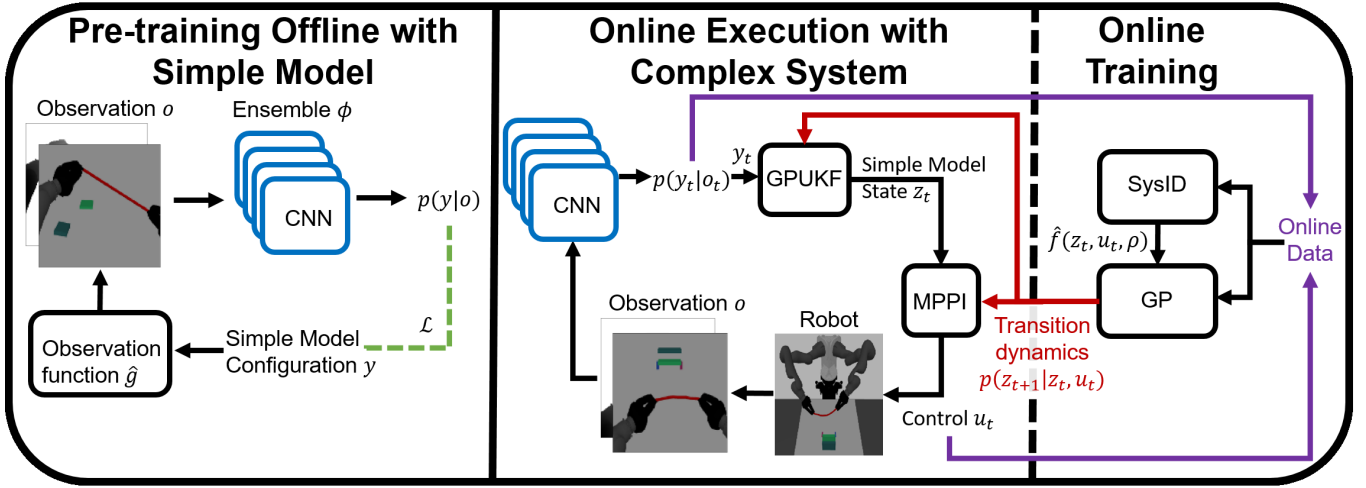} 
     \caption{Method overview. \textit{Left}: Training the CNN ensemble on image observations generated from the simple system offline. $\encoder$ is a CNN ensemble with variance used as a measure of uncertainty; \textit{Center}: Online execution using the simple model CNN with GPUKF filtering and MPPI for control; \textit{Right}: Procedure for fitting parameterized simple model and GP from observations of the complex system. The transition probability (red) is trained to predict the future uncertainty of $\encoder$, allowing us to avoid avoid areas where $\encoder$ is not confident.}%
     \label{fig:flow_diagram}
     \vspace{-0.6cm}
 \end{figure*}

Our full method is shown in Algorithm \ref{alg:main} and Figure \ref{fig:flow_diagram}. The overall procedure is to first generate a dataset of images with corresponding simple model configurations and then to train a perception system to estimate these configurations from images. Once this perception system is trained offline, we move to the online execution/learning phase, where we must manipulate the never-before-seen complex system.

The goal of the online execution is to reach a given goal region. However, because the perception system and the simple model dynamics can only account for \textit{some} complex model states, we must try to avoid states where the perception/dynamics are inaccurate. To this end, we collect data as we attempt the task and use that data to train a GP that captures the error in the simple model predictions. This error distribution is input into a Kalman Filter variant to better estimate the state and into a trajectory optimizer, which attempts to avoid regions of state space where the simple model predictions are inaccurate. The process of planning trajectories, executing one action, estimating the resulting state, and replanning a trajectory (Alg. \ref{alg:rollout}) repeats until the goal (or a timeout) is reached.


\setlength{\textfloatsep}{0pt}
\begin{algorithm}[t]
\caption{\methodname{}}
\textbf{Inputs:} Simple model dynamics: $\latentdynamics$;
Simple model cost: $\cost$;
Simple model renderer $\latentobservationfn$;
Initial data size $N$; $\#$ Episodes $K$
\begin{algorithmic}[1]
    \Algphase{Offline Training with simple system data}
    \State $\{y_i, o_i\}^N_{i=1} \gets$ \texttt{CollectData}($\latentdynamics, \latentobservationfn, N$);
    \State $\encoder \gets$ \texttt{TrainStateEstimator}($\{y_i, o_i\}^N_{i=1}$);
    \Algphase{Online Training with complex system data}
    \State $\mathcal{D} \gets \emptyset$; $\parameters, Q \gets$ \texttt{Initialize};
    \For{k $\in \{1, ..., K\}$}
        \State $p_z \gets \mathcal{N}(\latentdynamics(\latentstate_t, \control_t), Q(\latentstate_t, \control_t))$;
        \State $\mathcal{D} \gets \mathcal{D} \;\cup $; \texttt{Rollout}($p_z$, $\cost$, $\encoder$);
        \State $\parameters \gets $; \texttt{FitSimpleSystem}($\mathcal{D}$, $\latentdynamics$);
        \State $Q \gets $ \texttt{FitGP}($\mathcal{D}$, $\latentdynamics$, $Q$, $\parameters$);
    \EndFor
\end{algorithmic}
\label{alg:main}
\end{algorithm}

\subsection{Simple Model} \label{sec:simple_model}
The simple system state may contain elements which cannot be estimated from a single image, e.g. velocities. Thus we define the components of the simple state that can be noisily observed from a single image as latent observations $\latentobservation$. We then have the non-linear discrete-time state space model with dynamics described in Eq. (\ref{eq:dynamics}). In general there will be a non-linear mapping from $\latentstate$ to $\latentobservation$. In this paper we consider only a linear mapping, which is sufficient for our models:
\begin{equation}
\latentobservation_t = C \latentstate_t + \epsilon,
\label{eq:observation}
\end{equation}
For an $n$-dimensional simple model system ($\latentstate \in \mathbb{R}^n$) with $m$-dimensional ($m \leq n$) observations ($y \in \mathbb{R}^m$), $C = [I_{m \times m,}, 0_{m\times n-m}]$ selects the latent observations from $\latentstate$. For example, if $z$ is the position and velocity of a point, then $y$ is only the position, which is all that can be observed from a single image. In the case where $\epsilon \sim \mathcal{N}(0, R)$ for positive-definite $R$ we can use noisy measurements $\latentobservation$ to estimate $\latentstate$ by filtering using non-linear techniques such as the Unscented Kalman Filter (UKF) \cite{UKF}. We will show how to use a GP to learn $Q(z_t,u_t)$ in Eq. (\ref{eq:dynamics}) from data in Sec. \ref{sec:gp}.

\begin{algorithm}[t]
\caption{Rollout}
\textbf{Inputs:} Transition distribution   $p_z$;
Simple model cost: $\cost$;
CNN Ensemble $\encoder$
\begin{algorithmic}[1]
    \State $\mathcal{D} \gets \emptyset$; $\mu^z_1, \Sigma^z_1 \gets$ \texttt{Initialize};
    \For{t $\in \{1, ..., T\}$}
        \State $\mu^y_t, \Sigma^y_t \gets \encoder(o_t)$;
        \State $y_t \sim \mathcal{N}(\mu^y_t, \Sigma^y_t)$;
        \State $\mu^z_t, \Sigma^z_t  \gets $\texttt{GPUKF}($\mu^z_{t-1}, \Sigma^z_{t-1}, u_{t-1}, p_z,  y_t$);
        \State $u_t \gets$ \texttt{MPPI}$(\mu^z_t, c, p_z)$;
        \State $\mathcal{D} \gets \mathcal{D} \cup (\mu^y_t, \Sigma^y_t, u_t)$;
        \State \texttt{ExecuteAction($u_t$)};
        \If{\texttt{AtGoal}}
            break;
        \EndIf
    \EndFor
    \State \Return $\mathcal{D}$
\end{algorithmic}
\label{alg:rollout}
\end{algorithm}



\subsection{Probabilistic CNN Ensemble for Perception}\label{sec:CNN}
In order to use the simple model for the complex system, we need a perception system $\encoder$ that maps images to simple model states (even if the image is generated from the complex system). We would also like a way to estimate how well a simple model state approximates the complex system at a given state, as this gives us an estimate of confidence in the simple system dynamics at this state. We use the uncertainty in the perception estimate as a proxy for correspondence between the simple state and the unknown complex state. The perception output is
\begin{gather}    \mu^{\latentobservation}_t, \Sigma^{\latentobservation}_t = \encoder(\observation_t) \label{eq:perception}\\
    \latentobservation_t \sim p(\latentobservation_t | \observation_t) = \mathcal{N}(\mu^{\latentobservation}_t, \Sigma^{\latentobservation}_t),
\end{gather}
\noindent where the variance $\Sigma^\latentobservation_t$ estimates the uncertainty, and $\encoder$ is the perception system. We assume an isotropic Gaussian in Eq. \ref{eq:perception}, thus $\Sigma^\latentobservation_t$ can be described by a vector $\sigma^\latentobservation_t \in \mathbb{R}^m$. Ensembles have been empirically shown to give useful estimates of prediction uncertainty, which can be used to evaluate if a given input is out-of-distribution w.r.t the training data \cite{Ensembles}. Thus using ensembles avoids manually defining a similarity between the complex system observations and observations generated from the simple system. Instead we can input observation $\observation_t$ from the complex system into our perception system, and if it produces a high-certainty estimate of $\latentobservation_t$ (i.e. where $||\sigma^\latentobservation_t||$ is small), this implies that $y_t$ is a good approximation for the complex system at time $t$. 

\begin{figure}[t]
    \centering
    \includegraphics[width=.48\textwidth]{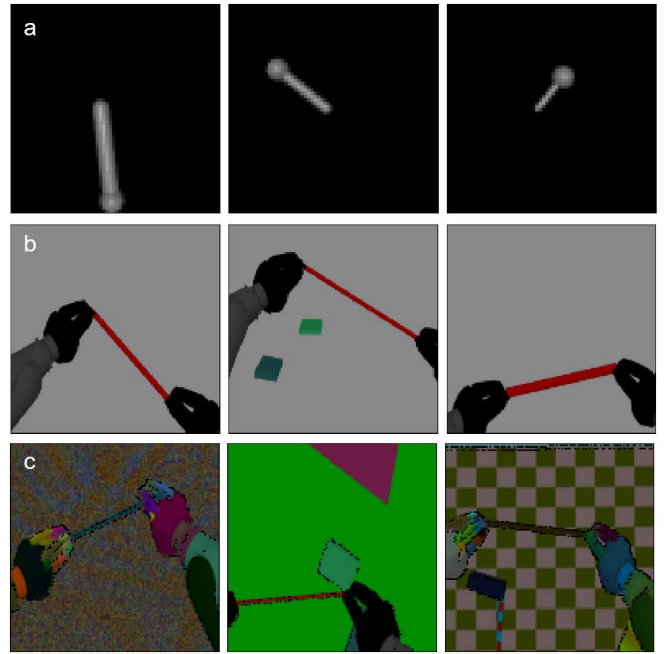}
    \caption{Examples of data generated from the simple system for training the CNN ensemble. (a) Tethered mass experiment, showing different geometries of the cart-pole. (b) Simulated rope manipulation experiment, showing different geometries of rigid link, and differing number and geometries of objects. (c) Real robot rope manipulation experiment. We randomize textures, lighting, obstacle configuration, camera pose, and rigid link geometry and add noise.
    }
    \label{fig:training_data}
\end{figure}

We parameterize $\encoder$ as a CNN ensemble which is trained with data generated from the simple system. Each CNN in the ensemble is a probabilistic CNN which outputs the parameters of a Gaussian, these are then combined into one Gaussian estimate. We train the CNN via supervised learning on observations of the simple system which we collect from simulation, along with correspond simple system states. Importantly, we assume that we can generate observations from the simple system which are similar to the complex system observations. To avoid requiring precise knowledge of the complex system before generating the simple model data, we generate a diverse training set of observations from the simple model. For example, we generate cart-poles with varying pendulum length for 
the tethered mass scenario. By generating diverse observations via domain randomization, our notion of visual similarity means that there is a simple system with some appearance and system parameters that looks similar to the complex system. See in Fig. \ref{fig:training_data} for examples.

Given an $\observation_t$ of the complex system online, we sample $y_t$ from the output of the $\encoder$ and use this along with the learned GP transition distribution (Sec. \ref{sec:gp}) to track a Gaussian distribution over the simple model state ($p(\latentstate_t | \control_{1:t-1}, \latentobservation_{1:t}) = \mathcal{N}(\mu^z_t, \sigma^z_t)$) with a GPUKF \cite{GPUKF}---an extension to the UKF for GP dynamics. When predicting $p(\latentstate_{t+1} | \control_{1:t}, \latentobservation_{1:t}$) in the GPUKF we use the posterior mean of the GP (Sec. \ref{sec:gp}) to perform the unscented transform, while the process noise is the posterior covariance of the GP, $Q(\latentstate_t, \control_t)$, evaluated at $(\mu^\latentstate_{t}, u_t)$.

\vspace{-5pt}
\subsection{System Identification} \label{sec:sysid}
The simple model dynamics may be parameterized by $\parameters$ (for example mass, length, etc.) and in order to use it, we must estimate the $\parameters$ which best approximates the complex system. One approach is using the Kalman filter to jointly estimate $\parameters$ and the latent state $\latentstate$, but we found that this was not numerically stable. Instead we use maximum-likelihood estimation on observed trajectories from the complex system. 

Given an observed trajectory of the complex system consisting of $\{\observation_t, \control_t\}^T_{t=1}$ we encode the observations into $\{\mu^{\latentobservation}_t, \sigma^{\latentobservation}_t, \control_t\}^T_{t=1}$. Since our trajectory may contain transitions which the simple model cannot accurately predict, we split the trajectory into $N$ trajectories of length $K < T$, and discard trajectories with average uncertainties above threshold $\alpha$ so we are left with high-certainty sub-trajectories. For each sub-trajectory we rollout the actions $\control_{1:T}$ using Eq. (\ref{eq:dynamics}) and (\ref{eq:observation}) to get estimated observations $\hat{\latentobservation}_{1:T}$ and perform gradient ascent on the parameters $\parameters$ and the trajectory initial states $\{\latentstate^i_1\}^N_{i=1}$ by maximizing the log likelihood of $\hat{\latentobservation}_{1:T}$ in the distribution output by the CNN ensemble $\mathcal{N}(\mu^{\latentobservation}_{1:T}, \sigma^\latentobservation_{1:T})$. The CNN weights are held constant. This process optimizes $\parameters$ to match the observed dynamics for high-certainty transitions in $(\latentstatespace, \controlspace)$.

\vspace{-5pt}
\subsection{Predicting Future Uncertainty with GP Regression} \label{sec:gp}
From $\encoder$ we have a confidence in our simple model approximation at a given $y$ (the uncertainty $\sigma^\latentobservation$). To keep the system in regimes where the approximation is accurate we also need to predict the future uncertainty conditioned on actions. Our uncertainty expresses uncertainty over the \textit{validity} of the state as a description of the complex system, rather than the \textit{value} of the state. Since we are using state uncertainty as a measure of confidence in the simple model approximation we model this uncertainty as state and action-dependent and use a GP with mean function $\latentdynamics$ and kernel function $\mathcal{K}$ to model the transition distribution. The GP posterior is
\begin{equation}
\label{eq:dynamics_2}
     p(z_{t+1} | z_t, u_t) =  \mathcal{N}(\latentdynamics(\latentstate_t, \control_t,\rho) + \mu_f(\latentstate_t, \control_t), Q(\latentstate_t, \control_t)),
\end{equation}
\noindent where $\mu_f$ and $Q$ are typically found via conditioning on some training set. However in our case this is a Gaussian Process State Space Model (GPSSM) \cite{VGPSSM} with transition probability above and emission probability defined in Eq. (\ref{eq:observation}). Training this GP is non-trivial as we do not have access to $\latentstate$ directly. Instead we must jointly infer both the transition probability and $\latentstate$ during training. 

We use a Parametric Predictive GP (PPGP)\cite{ParametricGPR} in order to train a GP with state-dependent aleatoric uncertainty via stochastic gradient descent. The uncertainty of the GP $\sigma^\latentstate$ is used to predict the uncertainty of the CNN ensemble $\sigma^\latentobservation$ via Eq. (\ref{eq:observation}). The PPGP is a sparse GP method which fits psuedo-inputs ($\zeta$) and psuedo-ouputs ($\gamma \sim \mathcal{N}(m, S)$) such that conditioning the GP on $(\gamma, \zeta$) approximates the true GP posterior. The GP parameters are thus $(m, S, \zeta)$ as well as the kernel hyper-parameters. The GP posterior contains an additional $\mu_f$ term compared with Eq. (\ref{eq:dynamics}). This allows the GP posterior mean to deviate from that of the simple model, attempting to fit transitions which do not conform to the simple model dynamics. Since our representation is known to be insufficient to model the true dynamics of the system, we are conservative and do not allow the GP to fit such transitions by constraining $m = 0$ and thus $\mu_f = 0$. We compare to a variant of our method where we do not enforce $\mu_f = 0$ in our experiments. 

We now describe how to train this GP using trajectories from the complex system of the form $\{\mu^{\latentobservation}_t, \sigma^{\latentobservation}_t, \control_t\}^T_{t=1}$. We would like Eq. (\ref{eq:observation}, \ref{eq:dynamics_2}) and an initial $p(z_1)$ to be able to reproduce the trajectory and uncertainties from the CNN. The learning objective to be minimized is then
\begin{equation}
    \mathcal{L} = \mathcal{KL}(p(\latentobservation_{1:T} | \observation_{1:T}) || p(\latentobservation_{1:T} | \control_{1:T})),
\end{equation}
\noindent where $\mathcal{KL}$ is the Kullback–Leibler divergence, $p(\latentobservation_{1:T})$ represents the joint distribution $p(\latentobservation_1, ..., \latentobservation_T)$, $p(\latentobservation_{1:T} | \observation_{1:T})$ is the output of the CNN, and  $p(\latentobservation_{1:T} | \control_{1:T})$ is the prediction from the dynamics and Eq. (\ref{eq:observation}). The GP predicted uncertainty $\sigma^\latentstate_t$ is used with  Eq. (\ref{eq:observation}) to predict a latent observation uncertainty $\hat{\sigma}^\latentobservation_t$. This objective aims to make the predicted uncertainty $\hat{\sigma}^\latentobservation_t$ and the observed uncertainties $\sigma^y_t$ consistent, i.e. the GP will predict the future uncertainty. 

$p(\latentobservation_{1:T} | \observation_{1:T})$ is fixed (i.e. we are not retraining the CNN online). Given this, we can rewrite the objective in terms of expectations over $p(\latentobservation_{1:T} | \observation_{1:T})$
\begin{equation}
    \mathcal{L} =  -\mathbb{E}_{p(\latentobservation_{1:T} | \observation_{1:T})}\left[ \log p(\latentobservation_{1:T} | \control_{1:T}) \right] + \mathcal{H} \left[ p(\latentobservation_{1:T} | \observation_{1:T}) \right]\label{eqn:objective_1},
\end{equation}
\noindent where $\mathcal{H}$ is the entropy and this entropy term can be dropped as it only depends on the pre-trained CNN. We can then optimize by maximizing the conditional expectation in Eq. (\ref{eqn:objective_1}) of $\latentobservation_{1:T}$. To do this we construct a variational lower bound on $p(\latentobservation_{1:T} | \control_{1:T})$.  This lower bound is given by
\begin{equation}
\begin{split}
    ELBO = \sum^T_{t=1}\mathbb{E}_{q(\latentstate_{t})} \left[\log p(\latentobservation_t | \latentstate_t) \right] - \mathcal{KL}(q(z_1) || p(z_1)) - \\
    \sum^T_{t=2} \mathbb{E}_{q(\latentstate_{t-1})} \left[ \mathcal{KL} \left( q(\latentstate_t) \;||\; p(\latentstate_t | \latentstate_{t-1}, \control_{t-1}) \right) \right],
\end{split}
\end{equation}
where the prior on the initial state is $p(z_1) \sim \mathcal{N}(0, I)$ and $q(z_t) = p(\latentstate_t | \latentobservation_{1:t}, \control_{1:t-1})$ is the GPUKF filtering distribution \cite{GPUKF}. The final objective to minimize is given by $\mathcal{L}_1$
\begin{equation}
    \mathcal{L}_1 =   -\mathbb{E}_{p(\latentobservation_{1:T} | \observation_{1:T})}[ELBO] \geq \mathcal{L} 
\end{equation}
To evaluate this objective we use the reparameterization trick to sample from the CNN and estimate gradients for $\mathcal{L}_1$. After performing this training procedure we obtain the transition distribution $p_z$, which is used by the GPUKF to perform filtering and by the MPC to predict future uncertainty.
\vspace{-5pt}
\subsection{Model Predictive Control} \label{sec:mpc}
For MPC we use MPPI \cite{MPPI} with a cost $\cost$ for the given task. To encourage the controller to keep the system in the domain of the simple model we add a cost to penalize the predicted uncertainty. Thus the cost function has the form $\cost(\latentstate, \sigma^\latentstate, \control)$ (examples are shown in the experiments). Note that typically in this setting the expected cost is computed, but as mentioned in the previous section, our uncertainty does not express uncertainty over the \textit{value} of the state. When rolling out a predicted trajectory with the model we propagate the expectation through the dynamics and record the one-step uncertainty for each step resulting in a trajectory $(\mu^z_t, \sigma^z_t, u_t)^T_{t=1}$ with which to calculate the cost. If we do not penalize this uncertainty, it will be ignored, which is equivalent to assuming the simple model is always accurate (we compare to this method in our experiments). Also, because we manually design the simple model state representation, we can incorporate additional information, such as avoiding collision, into the cost, which would have to be learned for an unsupervised learned representation.
\section{Experiments}
We evaluate \methodname{} on 1) manipulating a tethered mass, and 2) placing a rope in a narrow opening vs. baselines in the low-data regime. An episode is a time-limited attempt to reach the goal (terminating early when the goal is reached). See the accompanying video for example task executions.
\subsection{Environments}
\begin{figure*}[h]
    \centering
    \includegraphics[width=.99\textwidth]{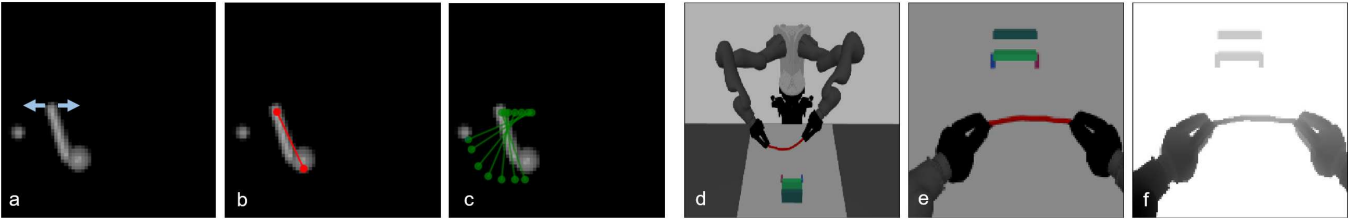}
    \caption{\textit{(a)} Tethered mass input image (64x64 grayscale) with the target (left) and the single prismatic joint (blue); \textit{(b)} output from CNN ensemble and GPUKF estimation (red); \textit{(c)} planned trajectory from MPPI (green). Only the first action from this trajectory is executed before replanning; \textit{(d)} The rope manipulation environment. 
    The goal is to bring the centre of the rope to the centre of the narrow gap. The sides of the gap have protrusions which can catch the rope; \textit{(e, f)} Example RGB and D observations from overhead Kinect.}
    \label{fig:results_combined}
    \vspace{-0.5cm}
\end{figure*}

\paragraph{Tethered Mass} This task involves controlling a tethered mass by applying force to the base of the tether. The goal is to bring the mass to a target without the tether contacting the target (tether contact results in failure). 
We implement this system in MuJoCo \cite{Mujoco}. There is a single actuated horizontal joint at the top of the tether (see Figure \ref{fig:results_combined}). 
Goals are randomly assigned at the start of each trial. This example demonstrates the applicability of \methodname{} to highly-dynamic systems where velocity must be considered.

The simple system we choose here is the pendulum on a cart (i.e. a cart-pole); we choose this because we observed that when the tether is taut the system will behave like a pendulum. We use an analytical dynamics function for $\latentdynamics$. We define $\latentstate = [p_{x_{cart}}, p_{x_{mass}}, p_{y_{mass}},\dot{p}_{x_{cart}}, \dot{\theta}]$, where $\theta$ is the angle of the pendulum. We define the latent observations as $y = [p_{x_{cart}}, p_{x_{mass}}, p_{y_{mass}}]$ and thus $C = [I_{3\times3} \;\; 0_{3\times2}]$. The parameters $\parameters$ are [\texttt{mass\_cart, mass\_pole, angular\_damping}]. 

\paragraph{Rope Manipulation} This task consists of two KUKA iiwa 7-DOF arms holding the ends of a rope. The goal to bring the center of the rope to the center of a narrow gap between two obstacles. These obstacles have small protrusions on which the rope can become caught. We implement this environment in Gazebo with the ode45 back-end (Figure \ref{fig:results_combined}). The action space of the robot is $[\Delta p_L, \Delta p_R] \in \mathbb{R}^6$ where $p_L, p_R$ are the left and right end-effector positions, respectively. We use a Jacobian-based method for inverse kinematics so that transitions in the robot's configuration space are smooth. The observations consists of RGBD data from an overheard Kinect. The goal and obstacle configuration for the task remain fixed across trials, but the starting locations of the end-effectors vary. We choose this example because it mimics cable installation, which is necessary for manufacturing and repair applications, where there are often narrow gaps and protrusions.

The simple system we choose here is to treat the rope as if it is a rigid link. The simple dynamics are then specified by adding a constraint that the gripper distances remain fixed. This approximation will be accurate so long as the rope is kept taut for the duration of the task. We define $z = y = [p_L, p_R]$ and $C = [I_{6 \times 6}]$. Since this model does not require dynamic parameters we forego the sysid step of our method.

\subsection{Baselines}
We compare \methodname{} to two recent methods from the literature. The first method is PlaNet \cite{PlaNet}, a model-based reinforcement learning algorithm. PlaNet learns a low-dimensional state representation along with dynamics and cost functions. The second is CURL \cite{CURL}, which uses a contrastive loss to learn a representation in which to learn a policy and has shown state-of-the-art sample-efficiency. For each of these baselines we test them by training them directly on the task with the complex system. We also show results for when the baselines are pre-trained on the simple system and fine-tuned on the complex system to investigate if these methods can take advantage of the data from the simple system. Both baselines were originally proposed with RGB observations, and we extend them to use RGBD for the rope experiment.

We also test with three variants of \methodname{}: 1) The full method which does both system identification and GP learning; 2) \methodname{} without the GP, 
this is equivalent to only using the simple model for control, and assuming it will be sufficiently accurate for all transitions. We choose this variant to investigate whether learning and avoiding inaccurate areas of the simple model state space is helpful for task performance; and 3) \methodname{} without constraining the GP posterior to be zero-mean, hence attempting to learn a better approximation of the dynamics in the simple system state space, rather than only where the simple model is accurate.
\subsection{Simple Model Data}
\paragraph{Tethered Mass}For pre-training the state estimator we generate 5000 trajectories of 20 time-steps from the cart-pole using random actions and render the cart-pole configurations to produce images. This corresponds to 100000  $64 \times 64$ grayscale frames. For domain randomization, we vary the dimensions and parameters of the system (see Figure \ref{fig:training_data}(a)).
\paragraph{Rope Manipulation}
For pre-training the state estimator we generate 800 trajectories of 50 time-steps length using random actions from the rigid body system and render the configuration. This corresponds to 80000 $128 \times 128 \times 4$ RGBD frames. For domain randomization, we vary the dimensions of the rigid link and the obstacles, as well as the obstacle locations (examples shown in Figure \ref{fig:training_data}(b)).
\subsection{Cost Functions}
For both LVSPC and PlaNet we use an MPC horizon of 40 and sample 1000 trajectories per timestep. We do not have a cost on control. CURL and PlaNet use the true environmental cost i.e. $c_{env}(x_t)$, whereas LVSPC and variants use an equivalent cost based on the simple model state with an uncertainty penalty $c(z_t, \sigma^z_t)$. The environmental costs use the true state from the simulator to calculate the cost  (because CURL and PlaNet have no knowledge of the simple model), whereas LVSPC uses the simple model state to approximate this cost, effectively giving CURL and PlaNet an advantage.
\paragraph{Tethered Mass}
The environmental cost consists of three parts; a euclidean distance to goal, a collision penalty for the tether and mass, and a penalty when the system goes out of view of the camera. The cost functions are $c(\latentstate_t, \sigma^{\latentstate}_t) = \delta_g \texttt{distToGoal}_{\latentstatespace} + \texttt{OffScreen}(\latentstate_t) + 10 \texttt{checkCollision}(\latentstate_t) +  \beta \sigma^\latentstate _t$ and $ c_{env}(\truestate_t) = \delta_g \texttt{distToGoal}_{\truestatespace} + \texttt{OffScreen}(\truestate_t) + 10 \texttt{checkCollision}(\truestate_t)$,
where $\beta$ is a parameter on how heavily to weigh uncertainty, and $\delta_g$ is $0$ if the goal is reached before time $t$ and $1$ otherwise. To balance exploiting vs. exploring we increase $\beta$ from $0$ to $2.0$ in the first 10 episodes. This cost is not memoryless; $\delta_g$ depends on the state for times $t' < t$. This is because we only wish to hit the target, we do not have to reach the target and stay there. 

\paragraph{Rope Manipulation}
The environmental cost is the distance to the goal, computed by considering 
the centre of the rope to be a floating point, discretizing the 3D environment into a 8-connected graph and solve for the shortest path to the goal for every point in the graph. We do not penalize contact for the baselines, as we found that they could exploit contact to help complete the task. The cost for \methodname{} penalizes contact (because the simple model is rigid), where we do a collision-check for the rigid-link approximation. The cost functions are  $c(\latentstate_t, \sigma^{\latentstate}_t) = \texttt{distToGoal}_\latentstatespace +  \beta \sigma^\latentstate _t + 100 \texttt{checkCollision}(\latentstate_t)$ and  $c_{\text{env}}(\truestate_t) = \texttt{distToGoal}_\truestatespace$.

To balance exploiting vs. exploring we increase $\beta$ from $0$ to $1000$ in the first 10 episodes. 

\subsection{Network Architectures}
Networks are implemented in PyTorch \cite{pytorch}, and the GPs are implemented in GPytorch \cite{gpytorch}, which allows us to exploit parallelism on the GPU for GP inference when performing MPPI. Thus, for the rope manipulation experiment, an iteration of MPPI takes only $0.89$s on average using an Intel i7-8700K CPU and an Nvidia 1080Ti GPU.
For both experiments we use a CNN ensemble consisting of 10 networks. All convolutional filters have filter size $3\times3$ and stride 2 for downsampling, all layers other than the output layers use ReLU activations. We use the Adam optimizer with a learning rate of $10^{-3}$, except when fine-tuning the pretrained CURL and PlaNet models where we use $10^{-4}$. 

For the GP dynamics model, we use 200 inducing points. We train an independent GP for each output dimension using the RBF kernel with automatic-relevance determination \cite{ARD}. We use a learning rate of $10^{-2}$ to train the GP and perform sysid. For each experiment CURL and PlaNet use encoders with the same architecture as our CNN. The transition and reward models for PlaNet are the same architecture as \cite{PlaNet}. The actor-critic architecture for CURL is the same as in \cite{CURL}. Both CURL and PlaNet are trained end-to-end.
\paragraph{Tethered Mass} 
Each CNN consists first of 4 convolutional layers. There is then a fully connected layer with 2048 hidden units, followed by an output layer.
\paragraph{Rope manipulation} 
Each CNN seperately processes depth and RGB, consisting of an RGB module and a depth module which are combined downstream. Each module consists first of 4 convolutional layers. There is then a fully-connected layer with 512 hidden units. After passing the RGB image through both the RGB module, and the depth image through the depth module, the output from each module is combined and passed through a final hidden layer of 1024 units, followed by an output layer.

\subsection{Results}
\begin{figure*}[h]
    \centering
    \includegraphics[width=.99\textwidth]{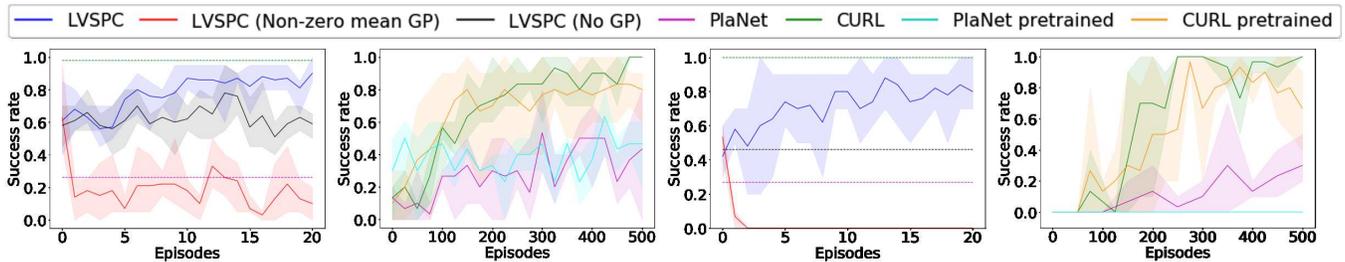}
    \caption{Average Success over 10 test tasks vs number of episodes for both experiments. Shaded region shows minimum and maximum success rate over 5 runs for \methodname{} and ablations and 3 runs for the baselines for a total of 50 and 30 test tasks for \methodname{} and the baselines, respectively. \textit{a)} \methodname{} and ablations for tethered mass, dotted lines show baseline performance after 500 episodes. \textit{b)} Baselines for tethered mass. \textit{c)} \methodname{} and ablations for rope, dotted lines show baseline performance after 500 episodes. \textit{d)} Baselines for rope.}
    \label{fig:statistical_results}
    \vspace{-0.5cm}
\end{figure*}
\paragraph{Tethered Mass} An example of the system tracking and MPC is demonstrated in Figure \ref{fig:results_combined}. Our statistical results are shown in Figure \ref{fig:statistical_results}(a, b). PlaNet achieves it's maximum performance at 200-300 episodes and has a success rate of approximately $26\%$ with large variation. We see that CURL shows the highest asymptotic performance, with $97\%$ after 400 episodes. Higher asymptotic performance is typical of model-free learning methods. Pre-training both PlaNet and CURL on data from the simple system results in improved initial performance, but lower final performance. In contrast, \methodname{} achieves approximately $90\%$ after 20 episodes, outperforming PlaNet and matching CURL's performance after 200 episodes, demonstrating 10x improved data efficiency. We also see that seeking to learn the dynamics in the simple state space with the GP results in substantially worse performance. This is likely because the simple state representation is insufficient to model the full complex dynamics. 
\paragraph{Simulated rope manipulation} Our statistical results are shown in Figure \ref{fig:statistical_results}(c, d). PlaNet's performance after 500 episodes is approximately $30\%$, while CURL solves the task with almost $100\%$ success rate after 250 episodes. Pre-training CURL on data from the simple system results in improved initial performance, but lower final performance, however pretraining PlaNet led to poor performance which it could not recover from, getting caught on the obstacles in every episode. Our full method achieves $80\%$ success rate after 20 episodes, again equivalent to CURL's performance after 200 episodes (thus we have 10x better data-efficiency) and outperforming PlaNet's final performance. We see that naively treating the rope as a rigid object results in approximately $46\%$ success and almost all failures result from the rope snagging on the protrusions on the side of the gap. As in the tethered mass experiment, attempting to fit the complex dynamics in the simple mode space is ineffective, causing frequent snagging on obstacles.

\subsection{Rope Manipulation on a Real Robot}
Our simulation experiments show that \methodname{} is effective at transferring within the same simulation environment. To validate that we can still use \methodname{} when the simple model and complex environments are very different, we perform the rope manipulation experiment described above on a real robot using a perception system trained only in simulation. 
We use domain randomization to improve the transfer of the CNN ensemble to real data  \cite{DomainRandomization} (see Figure \ref{fig:training_data}(c)). We observed better generalization when we randomized the pose of the camera and trained the CNN ensemble to produce an estimate in the camera frame instead of the world frame. 

We perform the experiment on the real robot over 5 random seeds. For each seed, after every 5 episodes
we record the success rate on 10 test episodes. The results are shown in Table \ref{table:real_robot_results}. Using \methodname{} we can complete this task with $80\%$ success using only 10 episodes of data collected on the real robot. This experiment demonstrates that using LVSPC is promising for real-world tasks, as we only need data from simulation to train an effective perception system.


\begin{table}[t]
\begin{center}
\begin{tabular}{ c|c|c|c|c|c } 
 Episode & 0 & 5 & 10 & 15 & 20 \\
 \hline
 Success rate& 0.3 & 0.7 & 0.8 & 0.78 & 0.82 \\ 
\end{tabular}
\caption{Results over 5 random seeds for real robot experiment}
\label{table:real_robot_results}
\end{center}
\end{table}

\section{Conclusion}
We have presented \methodname{} for leveraging a given simple model approximation to improve data efficiency for control tasks on systems with complex dynamics from image observations. We demonstrated this method on two tasks, showing substantially improved performance in the low-data regime over recent reinforcement learning methods. We have also demonstrated that we can apply our framework to a real robot while only using simulated data for pre-training. We assumed that the user specifies a type of simple model, but choosing a simple model which can approximate the complex system is an open problem, made difficult by the requirement that it must be possible to complete the task while operating only in the regime where the simple model is accurate. In future work we intend to incorporate multiple simple models and create a way to decide which is most appropriate. 
\bibliographystyle{IEEEtran}
\bibliography{references.bib}
\end{document}